\documentclass[letterpaper, 10 pt, conference]{ieeeconf}  

\usepackage[whole]{bxcjkjatype} 

\IEEEoverridecommandlockouts                              

\usepackage{graphicx}
\usepackage{epsfig} 
\usepackage{mathptmx} 
\usepackage{times} 
\usepackage{amsmath} 
\usepackage{amssymb}  

\usepackage{multirow}
\usepackage[table,xcdraw]{xcolor}
\usepackage{threeparttable}
\usepackage{eucal}
\usepackage{balance}
\usepackage{mathrsfs}  
\usepackage{comment}
\usepackage[noend]{algpseudocode}
\usepackage{float}
\usepackage{textcomp}
\usepackage{mathcomp}
\usepackage{bm}
\usepackage{cite} 
\usepackage{booktabs}
\usepackage{url}

\usepackage{hyperref}
\usepackage{stfloats}
\usepackage{capt-of}
\usepackage{eso-pic}

\def\eg{\textit{e.g}\onedot} 
\def\ie{\textit{i.e}\onedot}

\algnewcommand{\Break}{\textbf{break}}

\newcommand{\myvec}[1]{\mathbf{#1}}

\newcommand{\mytag}[1]{\texttt{#1}}

\newcommand{\higherbetter}{\raisebox{0.05ex}{\scriptsize$\uparrow$}}
\newcommand{\lowerbetter}{\raisebox{0.05ex}{\scriptsize$\downarrow$}}

\def\eg{{\it e.g.}}

\def\ie{{\it i.e.}}

\newboolean{showchanges}
\setboolean{showchanges}{true} 

\ifthenelse{\boolean{showchanges}}{
  
  \newcommand{\deleted}[1]{\textcolor{red}{\st{#1}}}
  
}{
  
  \newcommand{\deleted}[1]{}
  
}

\usepackage[colorinlistoftodos]{todonotes}

\newcommand{\preprintnotice}{%
  This work has been submitted to the IEEE for possible publication.
  Copyright may be transferred without notice, after which this version
  may no longer be accessible.}
\AddToShipoutPictureFG*{%
  \AtPageUpperLeft{%
    \put(\LenToUnit{0.5\paperwidth},\LenToUnit{-1.2cm}){%
      \makebox[0pt][c]{%
        \parbox{\textwidth}{\centering\footnotesize\preprintnotice}%
      }%
    }%
  }%
}

\title{\LARGE \bf
ReVNM: Learning-Based Visual Navigation from a Remote Camera
}

\author{Michikuni Eguchi$^{1,\dagger}$, Kohei Honda$^{2, 3,\dagger}$, Masafumi Endo$^{2}$, Yasuhiro Yoshimura$^{2}$, Ryo Yonetani$^{2}$
\thanks{$^{\dagger}$Equal contribution. }
\thanks{$^{1}$The Graduate School of Comprehensive Human Sciences, University of Tsukuba, Ibaraki, Japan. Work done during an internship at CyberAgent AI Lab. {\tt\small egrt117@gmail.com}}
\thanks{$^{2}$CyberAgent AI Lab, Tokyo, Japan, {\tt\small \{honda\_kohei, endo\_masafumi, yoshimura\_yasuhiro, yonetani\_ryo\}@cyberagent.co.jp}}%
\thanks{$^{3}$The Department of Mechanical Systems Engineering, Nagoya University, Aichi, Japan}%
\thanks{*This work was partially supported by Tokai Pathways to Global Excellence, part of MEXT Strategic Professional Development Program for Young Researchers. }
}

\newsavebox{\teaserbox}
\newlength{\teaserpad}
\makeatletter
\newcommand{\teaserblock}{%
  \sbox{\teaserbox}{\parbox[t]{\textwidth}{%
    \centering
    \includegraphics[width=\textwidth]{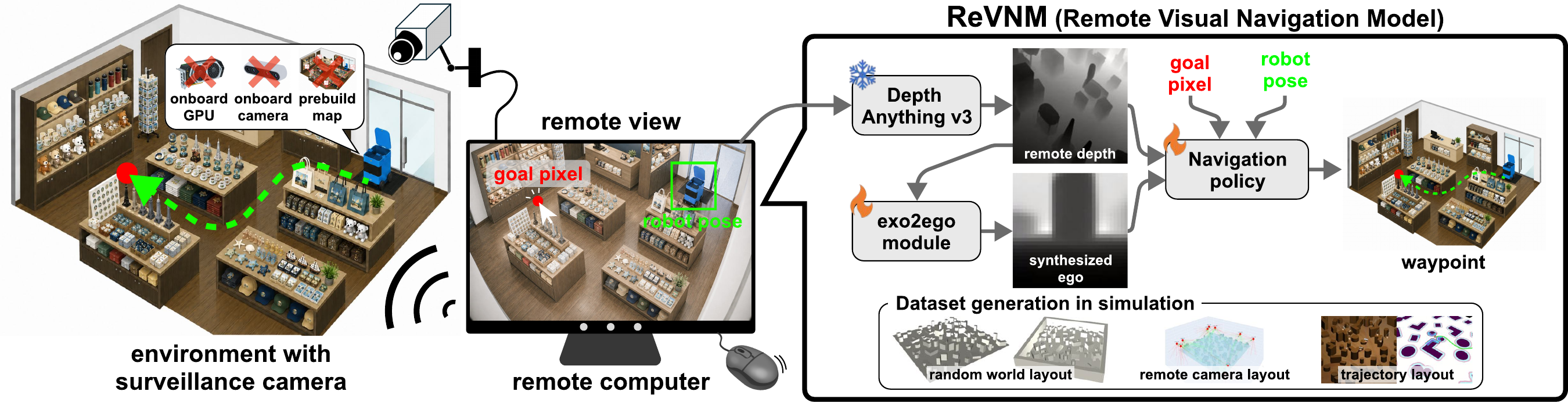}%
    \captionof{figure}{ReVNM enables robot navigation with a single uncalibrated exocentric camera and requires no onboard vision processing or a pre-built map. The core of ReVNM, an exocentric-to-egocentric (exo2ego) module, synthesizes egocentric depth from the remote image. The synthesized depth fills in the robot's surroundings that the remote camera alone cannot capture and helps the navigation policy predict waypoints.}%
    \label{fig:teaser}}}%
  \@tempdima=\baselineskip
  \@tempdimb=\dimexpr\ht\teaserbox+\dp\teaserbox\relax
  \@tempcnta=\@tempdimb \divide\@tempcnta by \@tempdima \advance\@tempcnta\tw@
  \setlength{\teaserpad}{\dimexpr\@tempdima*\@tempcnta\relax}%
  \setbox\teaserbox=\vbox to\teaserpad{\vfil\box\teaserbox\vfil}%
  \nointerlineskip
  \box\teaserbox
  \nointerlineskip
}
\makeatother

\IEEEaftertitletext{\vspace{-2\baselineskip}\teaserblock}

\begin{document}

\bstctlcite{BSTcontrol}

\maketitle

\thispagestyle{empty}
\pagestyle{empty}


\begin{abstract}
Visual Navigation Models (VNMs) enable robots to navigate from egocentric visual observations without geometric localization and planning, but long-range navigation still requires pre-built maps. This paper presents the Remote Visual Navigation Model (ReVNM), which uses a single remote surveillance camera to serve as both an observation source and an implicit environmental map for visual navigation.
While the use of remote cameras could eliminate the need for pre-built maps as well as onboard vision processing, their limited field of view instead of egocentric observations makes it hard to achieve collision-free navigation. The lack of existing data with diverse remote viewpoints, which are crucial for training robust VNMs, further complicates the challenge.
In this work, we propose a learning-by-synthesis approach to address this two-fold challenge. Our ReVNM extends a state-of-the-art VNM architecture with an exocentric-to-egocentric (exo2ego) module that predicts an egocentric depth observation from remote-camera observations. This helps the VNM to plan a path while considering obstacles in front of the robot. Trained only on randomly generated worlds with diverse obstacle layouts and camera viewpoints, ReVNM can generalize well to real robot navigation without additional fine-tuning. 
Experiments in both simulation and real-world environments confirmed the effectiveness of the proposed approach.

\end{abstract}


\section{INTRODUCTION}
\label{sec:introduction}

Visual navigation models (VNMs) are imitation-learned policies that guide a robot to a goal from egocentric visual observations alone, without explicit geometric localization or path planning. Existing models such as GNM~\cite{shah2023gnm}, ViNT~\cite{shah2023vint}, and NoMaD~\cite{NoMaD} have demonstrated strong generalization across environments and robot platforms.

However, deploying a VNM entails two practical requirements. First, navigation using VNMs, especially when the goal is outside the initial view of the robot, typically relies on a pre-built map that must be updated as the environment changes. Second, running the model in real time requires onboard cameras and sufficient GPU resources, which limits its applicability to low-cost robots. These requirements motivate our central question: \emph{can we use VNMs without pre-built mapping and compute-intensive onboard visual processing?}

In this work, we propose the \emph{Remote Visual Navigation Model (ReVNM)}. 
Our key idea is to use a fixed remote surveillance camera as a persistent, map-like observation source.
Compared to mobile robots, surveillance cameras have already been deployed widely in real-world environments. Their continuous recording is a natural map of the environment that is always up to date. These cameras are often connected to network compute resources, to which robots can delegate intensive visual processing by VNMs.

While intuitive, the use of remote camera vision for VNMs comes with unique challenges. Although an exocentric viewpoint provides a broader view of the environment, obstacles can occlude the robot's immediate surroundings, particularly under oblique camera viewpoints. As a result, the remote observation alone may not contain sufficient local geometric information for collision-aware waypoint prediction.
Another challenge is the lack of existing data compared to the conventional egocentric-camera setup~(\eg, \cite{shah2023gnm}). Although surveillance cameras are ubiquitous, most of them are proprietary, and collecting diverse data for training robust models is impractical.

The proposed ReVNM employs a learning-by-synthesis approach to address these challenges (Fig.~\ref{fig:teaser}). Specifically, we construct a large-scale synthetic dataset comprising expert demonstrations of mobile robot navigation in randomly generated worlds, recorded as exocentric videos from diverse viewpoints along with their egocentric counterparts. Rather than relying solely on exocentric video feeds, ReVNM learns an exocentric-to-egocentric (exo2ego) diffusion module that predicts egocentric depth observations from the exocentric inputs. This enables the robot to detour around predicted obstacles ahead and to identify narrow gaps for effective shortcuts. Moreover, we introduce a data augmentation strategy inspired by DAgger~\cite{dagger}, which incorporates recovery demonstrations collected when the robot reaches near-collision states that the expert trajectories rarely visit. By integrating these techniques, we show that ReVNM can effectively generalize to unseen environments, including real-world cluttered offices.

In summary, our contributions are as follows:
\begin{itemize}
\item We formulate a remote-camera visual navigation setting in which a robot navigates to an image-specified goal using a single uncalibrated exocentric camera, without requiring a pre-built global map or onboard vision processing.

\item We propose ReVNM, which combines remote depth observations with an exo2ego diffusion module that predicts complementary egocentric depth for robot-centric waypoint prediction. The model is trained entirely from synthetic expert demonstrations together with DAgger-inspired recovery data.

\item We evaluate ReVNM in diverse simulated environments and on a different real robot, demonstrating improved navigation performance over the image-space remote-navigation~\cite{robinson2023visual} and egocentric-VNM~\cite{NoMaD} baselines.
\end{itemize}

\section{RELATED WORK}
\label{sec:related_work}
\subsection{Visual Navigation Models}
VNMs are goal-conditioned policies that predict short-horizon waypoints from visual observations, typically captured by onboard cameras. When trained on large-scale egocentric video datasets, VNMs have demonstrated strong generalization across environments and robot platforms~\cite{shah2023gnm,shah2023vint,NoMaD}. For long-range navigation beyond a single egocentric view, they are typically combined with topological maps~\cite{savinov2018semi}, whose nodes store egocentric observations and pre-computed traversability between nodes.

Recent work has explored richer environment representations beyond topological maps. 
For example, 3D scene graphs~\cite{garg2025objectreact}, Gaussian splatting~\cite{honda2025gsplatvnm}, and video-generative world model~\cite{bar2025navigation} are used as environmental maps for VNMs. 

In this work, we seek another direction, \ie, using remote camera observations as input to VNMs. This not only mitigates the problem of the limited egocentric field of view for long-range navigation, but is also advantageous for low-cost robot applications where onboard vision and compute infrastructure is not always affordable.

\subsection{Model-Based Remote Visual Navigation}
Navigation from a fixed remote camera, while it is new in the context of VNMs, has traditionally been studied through model-based visual servoing~\cite{wu2022survey}.
Position-based visual servoing (PBVS) reconstructs the robot, goal, and obstacles in a physical coordinate frame, where conventional localization~\cite{janabi2010kalman,lippiello2004visual,bultmann2023external} and control~\cite{dixon2001adaptive} methods can be applied. Its performance, however, depends on reliable geometric reconstruction, which is difficult with unknown camera parameters or partial occlusion~\cite{yang2023bevheight}.
Image-based visual servoing (IBVS) instead operates directly in image space, typically using a model that relates control inputs to image-space motion. Starting from \cite{rao2005planning}, existing works have studied adaptive control~\cite{liang2015adaptive}, uncalibrated cameras~\cite{liang2020purely,robinson2026robot}, or robot pose estimation~\cite{robinson2023visual}.

A fundamental difficulty in model-based IBVS is that the relationship between image displacement and physical robot motion strongly depends on the camera viewpoint and image location, particularly for oblique cameras. Rather than explicitly deriving this mapping, ReVNM learns it from demonstrations by directly predicting robot-centric waypoints from exocentric visual observations.

\section{ReVNM: Remote Visual Navigation Model}
\label{sec:method}

\subsection{Problem Formulation}
\label{subsec:task}

We consider a scenario with a fixed remote camera installed in the environment to monitor the entire space in which a robot navigates. Although the traversable space can be partly occluded by obstacles, we still assume that the initial position of the robot and its navigation goal are clearly visible from the camera. Formally, let $\myvec{o}_t$ be the observation at time $t$ and $\myvec{p}_\textrm{goal}$ be the goal. The task of visual navigation is to learn a navigation policy that predicts the conditional probability of the sequence of $H$ future waypoints from $t+1$, $\myvec{w}_{t+1:t+H}$, as follows:
\begin{equation}
\pi(\myvec{w}_{t+1:t+H} \mid \myvec{o}_t, \myvec{p}_\textrm{goal}).\label{eq:task}
\end{equation}
We assume that this policy can be executed on the compute infrastructure attached to the remote camera. The robot is equipped with the minimum set of sensors (\eg, 2D LiDAR) and computers required for local motion planning to follow the planned path $\myvec{w}_{t+1:t+H}$. Each waypoint is represented in a robot-centric 2D coordinate frame in metric units.
With this setup, the policy infers these waypoints using information available from the remote camera, 
without relying on a pre-built map or onboard visual observations.

Note that this setting shares a structural analogy with exocentric visuomotor manipulation, where an external camera observes both the end-effector and its target.
In such settings, exocentric and egocentric views provide complementary global and local visual information~\cite{hsuvision}. Nevertheless, our problem assumes no onboard cameras available on the robot, motivating our solution to \emph{synthesize} the egocentric observation from available information instead.

\begin{figure*}[t]
    \centering
    \includegraphics[width=1.0\linewidth]{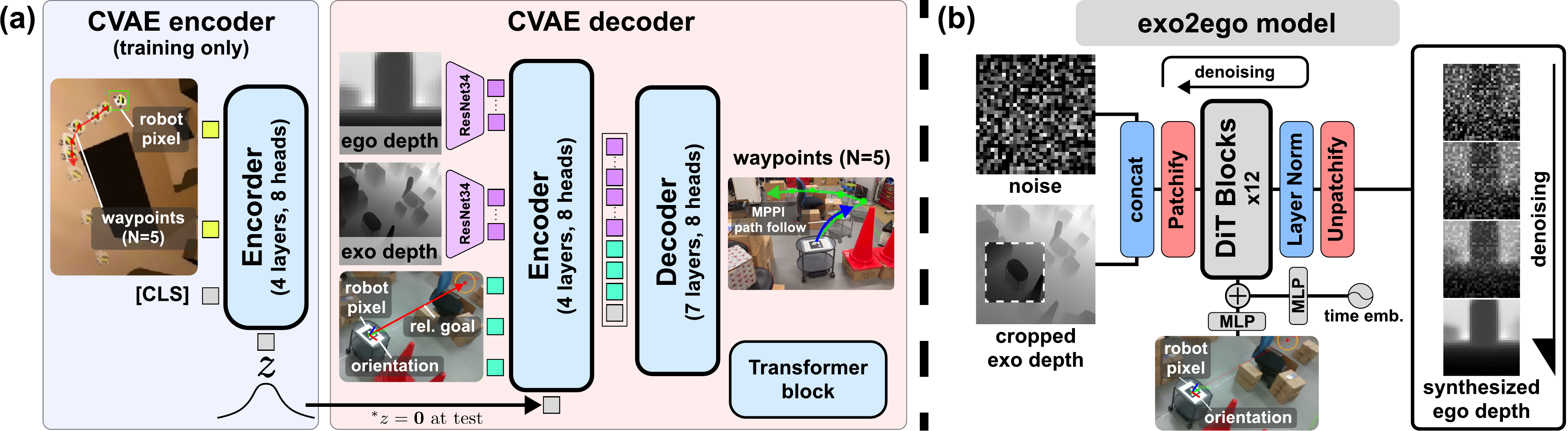}
    \caption{Model architecture of ReVNM. \textbf{(a)} The navigation policy is an ACT-style CVAE. It encodes the ego and exo depth with ResNet-34 and predicts a chunk of waypoints from these, the robot pixel, orientation, and goal. \textbf{(b)} exo2ego is a Diffusion Transformer that synthesizes the ego depth from the cropped exo depth, conditioned on the robot pixel and orientation.}
    \label{fig:model}
\end{figure*}

\subsection{Observation Model}
\label{subsec:observation}

We design the policy observation $\myvec{o}_t$ to provide the geometric information required for waypoint prediction while minimizing appearance-dependent cues that may hinder policy transfer across environments.
Specifically, the observation at time $t$ is
$
\myvec{o}_t = 
\left(
\myvec{d}^{\text{exo}}_{t-N+1:t},
\myvec{d}^{\text{exo2ego}}_{t-N+1:t},
\myvec{p}_{\text{rob}},
\myvec{q}_{\text{rel}}
\right),
$
where $\myvec{d}^{\text{exo}}_{t-N+1:t}$ denotes normalized exocentric depth images over the most recent $N$ time steps, obtained from the remote-camera RGB images using monocular depth estimation~\cite{lin2026depth}.
$\myvec{p}_{\text{rob}}$ and $\myvec{q}_{\text{rel}}$ denote the robot's normalized pixel location and relative orientation with respect to the remote-camera image coordinate frame, respectively.

One of our technical contributions is $\myvec{d}^{\text{exo2ego}}$, an egocentric depth observation generated from $\myvec{d}^{\text{exo}}_{t-N+1:t}$. This additional input is expected to better capture obstacles that may be occluded from the camera's exocentric viewpoint, as will be detailed in Sec.~\ref{subsec:exo2ego}.

\subsection{Policy Network Architecture}
\label{subsec:policy}

As shown in Fig.~\ref{fig:model}(a), we design the policy as a conditional variational autoencoder (CVAE)~\cite{CVAE} following ACT~\cite{act}, treating a sequence of $H$ future waypoints as an action chunk.
During training, a transformer-based encoder infers a latent variable $\myvec{z}$ from the robot location $\myvec{p}_{\text{rob}}$ and the target waypoints.
At test time, $\myvec{z}$ is fixed to the prior mean.
The policy decoder consists of a transformer~\cite{vaswani2017attention} encoder-decoder conditioned on $\myvec{z}$, the observation $\myvec{o}_t$, and the goal $\myvec{p}_{\text{goal}}$.
The transformer encoder fuses these inputs into a sequence of tokens: each depth image is encoded using a ResNet~\cite{ResNet} backbone with positional encodings, while $\myvec{p}_{\text{rob}}$, $\myvec{q}_{\text{rel}}$, $\myvec{p}_{\text{goal}}$, and $\myvec{z}$ are individually projected by linear layers.
The transformer decoder takes $H$ learned queries and attends to the encoded tokens to predict the future waypoints $\myvec{w}_{t+1:t+H}$ as an action chunk.

\subsection{Exocentric-to-Egocentric (exo2ego) Module}
\label{subsec:exo2ego}


The exo2ego module synthesizes the egocentric depth observation $\myvec{d}^{\text{exo2ego}}_t$ from the exocentric observation. 
We adopt a diffusion-based method that generates egocentric images for manipulation~\cite{spisak2024diffusingelsesshoesrobotic} to our remote visual navigation setting.
Unlike manipulation where the viewpoint is fixed, navigation involves an egocentric viewpoint that changes substantially as the robot moves. We therefore condition the exo2ego module not only on the exocentric depth but also on the robot pixel $\myvec{p}_{\text{rob}}$ and orientation $\myvec{q}_{\text{rel}}$.
Formally, the exo2ego module $g_\psi$ generates the ego depth as
\begin{equation}
    \myvec{d}^{\text{exo2ego}}_t =
    g_\psi(
    \tilde{\myvec{d}}^{\text{exo}}_{t-N+1:t},
    \myvec{p}_{\text{rob}},
    \myvec{q}_{\text{rel}}
    ),
    \label{eq:exo2ego}
\end{equation}
where $\tilde{\myvec{d}}^{\text{exo}}$ denotes the exocentric depth cropped around $\myvec{p}_{\text{rob}}$ to emphasize the robot's local surroundings, as shown in Fig.~\ref{fig:model}(b).

We model $g_\psi$ as a conditional denoising diffusion probabilistic model (DDPM)~\cite{ho2020denoising}, whose denoiser is a Diffusion Transformer (DiT)~\cite{peebles2023scalable}.
The denoiser predicts the clean ego-depth image from a noisy ego-depth sample, conditioned on the cropped exocentric-depth history and the robot pose, following the $x_0$-prediction formulation~\cite{he2025lotusdiffusionbasedvisualfoundation}.
The noisy ego-depth sample and the cropped exocentric-depth history are concatenated channel-wise, while the diffusion step and robot pose are used as conditioning inputs to the DiT.

\subsection{Training}
\label{subsec:training}
The navigation policy and the exo2ego module are trained separately using the same simulation dataset described in Sec.~\ref{sec:dataset}, with no gradient propagation between them.

\paragraph{Navigation Policy}
We train the ReVNM navigation policy, as described in Sec.~\ref{subsec:policy}, with an L1 reconstruction loss and a KL regularization term, as in~\cite{act},
\begin{equation}
    \mathcal{L} = \left\lVert \myvec{w}^{*}_{t+1:t+H} - \myvec{w}_{t+1:t+H} \right\rVert_1
    + \beta \, D_{\mathrm{KL}}\!\left( q_\phi \,\middle\|\, \mathcal{N}(\myvec{0}, \myvec{I}) \right),
    \label{eq:policy_loss}
\end{equation}
where $\myvec{w}^{*}_{t+1:t+H}$ are the expert waypoints of Sec.~\ref{sec:dataset}, $q_\phi = q_\phi(\myvec{z} \mid \myvec{p}_{\text{rob}}, \myvec{w}^{*}_{t+1:t+H})$ is the CVAE encoder, and $\beta$ weights the regularization.

\paragraph{exo2ego Module}
We pretrain the exo2ego module using a pixel-wise mean-squared error between the predicted depth and the ground-truth ego depth $\myvec{d}^{\text{ego}}_t$, obtained from the robot-mounted camera in simulation.

\section{Training Dataset for ReVNM}
\label{sec:dataset}
\subsection{Data Synthesis}




To address the lack of diverse exocentric visual navigation data, we adopt a learning-by-synthesis approach and construct a large-scale synthetic dataset of expert navigation demonstrations.
We automatically collect expert robot trajectories using a classical navigation planner and record synchronized exocentric and egocentric image sequences along each trajectory.
The trajectories provide expert waypoints $\myvec{w}^*$ for the navigation policy, while the paired exocentric and egocentric observations are used to train the policy and the exo2ego module.

To promote generalization to unseen environments, we generate a wide variety of navigation scenes by randomizing the geometric properties of obstacles rather than relying on a fixed set of hand-designed environments.
Specifically, we procedurally construct environments from simple geometric primitives with randomized sizes, shapes, densities, and layouts, including both pillar- and wall-like obstacles, as shown in Fig.~\ref{fig:teaser}.
We additionally randomize the exocentric camera configuration across demonstrations, exposing the model to diverse viewing positions and angles.
Together, these variations provide broad coverage of the scene geometry and camera configurations encountered in exocentric navigation.

\subsection{Data Augmentation}



Although classical navigation planners allow us to automatically collect many expert demonstrations, the resulting trajectories tend to be overly well behaved.
The expert typically follows near-optimal paths while maintaining clearance from obstacles, and therefore rarely visits states in which the robot is close to collision or must recover from navigation errors.
As we show in Sec.~\ref{subsec:dagger-effect}, training only on such demonstrations substantially reduces navigation robustness.

To complement these demonstrations, we augment the dataset using a DAgger-inspired~\cite{dagger} iterative recovery data collection procedure.
We roll out the current navigation policy and collect additional expert recovery trajectories from near-failure states encountered during navigation, such as when the robot approaches an obstacle or becomes stuck.
These recovery demonstrations are added to the training set, and the policy is retrained on the augmented dataset.

\section{EXPERIMENTS}
\label{sec:experiment}

We evaluate the navigation performance of ReVNM in both simulation and the real world. We compare ReVNM with an existing egocentric VNM and a model-based remote navigation method, analyze the contribution of the proposed components through ablation studies, and evaluate whether the simulation-trained model generalizes to a real robot without further training.

\subsection{Experimental Setup}
\label{subsec:setup}

\subsubsection{Simulation Setup}

Simulation experiments are conducted in Gazebo~\cite{koenig2004gazebo} using a differential-drive robot, Megarover (Vstone Co., Ltd.). The robot is equipped with a 2D LiDAR and an RGB camera ($640\times480$, $120^\circ$ FOV), and its maximum linear and angular velocities are $0.5$~m/s and $0.7$~rad/s, respectively. The same robot platform is used to generate the training data. The onboard RGB camera is used for training-data generation and simulation-only baselines, but is not used by ReVNM at inference.

We evaluate in the three environments shown in Fig.~\ref{fig:sim-world-result}. \mytag{Random Pillar} is generated using the same procedural environment generator as the training environments, while the evaluation uses held-out obstacle layouts that are not included in the training data. \mytag{Book Store} and \mytag{Warehouse} are realistic environments used only for evaluation. The former represents a retail store cluttered with tall shelves, while the latter is a large warehouse containing narrow passages and multiple dead ends.\footnote{\url{https://github.com/aws-robotics}} The remote camera has a resolution of $640\times480$ and a $120^\circ$ FOV, and its pose is varied across evaluation trials.

\subsubsection{Real-World Setup}

For the real-world experiments, we use a Kachaka (Preferred Robotics, Inc.), a differential-drive robot different from the platform used for training-data generation. The robot is equipped with a 2D LiDAR, with maximum linear and angular velocities of $0.5$~m/s and $1.9$~rad/s, respectively. No onboard RGB camera is used.

A RealSense D435 (Intel Corp.) is installed as the remote camera in an office environment. We evaluate two obstacle configurations (Fig.~\ref{fig:real-world-result}) designed to require different navigation behaviors. In \mytag{Forest}, five small obstacles are scattered across the workspace, requiring the robot to traverse narrow gaps between them. In \mytag{Wall}, a wall-like obstacle separates the start and goal and contains a narrow slit through which the robot can pass.

\subsubsection{Evaluation Protocol and Metrics}
\label{sssec:evaluation}

The robot performs the remote visual navigation task defined in Sec.~\ref{subsec:task}, where the goal is specified as a pixel coordinate in the remote-camera image. The robot's image-space position and orientation are available to the remote navigation system. In simulation, they are obtained from the simulator ground truth; when the robot is occluded, the last observed pose is retained. In the real world, they are obtained by detecting an AprilTag~\cite{olson2011apriltag} mounted on the robot.

All methods share the same LiDAR-based MPPI controller~\cite{williams2017mppi}, which tracks the generated waypoints while performing reactive local obstacle avoidance. The controller does not perform global planning, \ie, the overall direction toward the goal and global obstacle detours are determined by the waypoints generated by each navigation method.

A trial is considered successful if the robot reaches the goal within $60$~s without collision. A collision or failure to reach the goal within the time limit is counted as a failure. In the real-world experiments, leaving the remote camera's field of view without recovery is also counted as a failure.

In simulation, goal arrival is determined by the metric distance to the goal. The threshold is $0.5$~m for \mytag{Random Pillar} and \mytag{Book Store}, and $1.0$~m for the larger \mytag{Warehouse}. In the real world, goal arrival is determined by the image-space distance between the robot and the goal, with a threshold of $20$ pixels.

For each simulated environment, we evaluate 10 start--goal pairs under five remote-camera poses and repeat each condition five times, resulting in 250 trials per environment. The camera poses are sampled using the ray-casting procedure to ensure sufficient visibility of the workspace. The start and goal positions are randomly sampled from free space with a separation of 5--10~m.

For the real-world experiments, the remote camera is fixed. We conduct 25 trials in \mytag{Forest} (five start--goal pairs, five runs each) and five trials in \mytag{Wall} (one pair, five runs).

We report success rate (SR) in both simulation and the real world, and Success weighted by Path Length (SPL)~\cite{SPL} in simulation to evaluate path efficiency. SR is the fraction of successful trials. For SPL, we use the path planned by ROS Navigation 2 (Nav2)~\cite{macenski2020nav2} as the reference path instead of the geodesic shortest path, since the Nav2 path represents a feasible trajectory that the robot can actually follow.

\subsubsection{Baseline Methods}
\label{sssec:methods}

We compare ReVNM with two complementary baselines representing the two major alternatives to our approach: an egocentric VNM and a model-based image-space remote-navigation method. We additionally evaluate two ReVNM variants to isolate the contribution of the synthesized egocentric observation.

\begin{itemize}

\item \textbf{IBVS}: We construct an obstacle-aware image-space navigation baseline inspired by~\cite{robinson2023visual}. SAM3~\cite{carion2026sam3segmentconcepts} segments traversable regions in the remote image, and A* replans an image-space path every 1~s. Pure pursuit converts the planned image-space path into waypoints, which are tracked by the same LiDAR-based MPPI controller used by ReVNM. If no feasible A* path is found, the robot directly pursues the goal. Like ReVNM, this baseline requires neither an onboard RGB camera nor a pre-built map\footnote{We do not construct a PBVS baseline that unprojects the monocular depth into a ground-plane map. Recovering metric scale and a ground plane from a single uncalibrated oblique view is unreliable in cluttered scenes~\cite{yang2023bevheight}, and such a map would leave camera-occluded regions unobserved.}.

\item \textbf{NoMaD}~\cite{NoMaD}: NoMaD is a representative egocentric, map-based VNM baseline. Following its standard long-range navigation setting, we first drive the robot along the Nav2 reference route and record an egocentric image sequence, which is provided as the topological map. We evaluate both the original pretrained model (\textbf{w/o FT}) and a model fine-tuned on our synthetic training data (\textbf{w/ FT}). Because NoMaD requires onboard RGB observations, it is evaluated only in simulation.

\item \textbf{Ours w/o exo2ego}: An ablation of ReVNM that removes the synthesized egocentric depth and predicts waypoints using only the exocentric observation.

\item \textbf{Ours w/ oracle-ego}: A reference variant that replaces the synthesized egocentric depth with the ground-truth egocentric depth obtained from the onboard camera. This variant provides an upper-bound reference for the benefit of egocentric observations and is evaluated only in simulation.

\end{itemize}

Accordingly, the real-world experiments compare \textbf{IBVS}, \textbf{Ours w/o exo2ego}, and \textbf{Ours}.

\subsubsection{Implementation Details}
\label{sssec:implementation}

\paragraph{Model Configuration}
The navigation policy takes $224\times224$ egocentric and exocentric depth images from the most recent $N=5$ time steps and predicts a chunk of $H=5$ waypoints in approximately $0.1$~s. Each depth image is encoded by a ResNet-34~\cite{ResNet} pretrained on ImageNet-1k and fine-tuned during training. The navigation policy has 63.6M parameters.

The exo2ego module generates a $32\times32$ egocentric depth image, which is resized to $224\times224$ before being passed to the navigation policy. Its denoiser is a DiT-S/2~\cite{peebles2023scalable} with 32.8M parameters. At inference time, the module starts from Gaussian noise and generates one depth image in approximately $30$~ms with eight sampling steps of Denoising Diffusion Implicit Models (DDIM)~\cite{song2021denoising}.

\paragraph{Training}
The navigation policy is trained for 20 epochs using AdamW with a learning rate of $2\times10^{-4}$ and a weight decay of $10^{-4}$. We use the L1 waypoint loss and a KL regularization term with $\beta=1.0$. The exo2ego module is trained for 30 epochs using Adam with a learning rate of $1\times10^{-4}$. Both models use a cosine learning-rate schedule and a batch size of 512.

Following the data-generation procedure in Sec.~\ref{sec:dataset}, we collect approximately 50k episodes with Nav2 as the expert, which are divided into five-step segments to obtain approximately 1M training samples. Among them, approximately 5k episodes are recovery trajectories collected over two DAgger iterations.

\subsection{Experimental Results}
\label{subsec:results}

\subsubsection{Comparison with Baseline Methods in Simulation}
\label{sssec:result_sr}

\begin{table}[t]
  \centering
  \caption{Navigation performance in simulation.}
  \label{tab:sim_results}
  \setlength{\tabcolsep}{2.2pt}
  \renewcommand{\arraystretch}{0.95}
  \resizebox{\columnwidth}{!}{%
  \begin{tabular}{lcccccc}
    \toprule
    \multirow{2}{*}{Method}
      & \multicolumn{2}{c}{\mytag{Random Pillar}}
      & \multicolumn{2}{c}{\mytag{Book Store}}
      & \multicolumn{2}{c}{\mytag{Warehouse}} \\
    \cmidrule(lr){2-3} \cmidrule(lr){4-5} \cmidrule(lr){6-7}
      & SR\higherbetter & SPL\higherbetter & SR\higherbetter & SPL\higherbetter & SR\higherbetter & SPL\higherbetter \\
    \midrule
    NoMaD w/o FT
      & 5$\pm$3\% & 0.02$\pm$0.01
      & 18$\pm$5\% & 0.08$\pm$0.02
      & 20$\pm$5\% & 0.08$\pm$0.02 \\
    NoMaD w/ FT
      & 40$\pm$3\% & 0.33$\pm$0.03
      & 22$\pm$3\% & 0.09$\pm$0.02
      & 22$\pm$4\% & 0.09$\pm$0.02 \\
    IBVS
      & 78$\pm$2\% & 0.61$\pm$0.05
      & 75$\pm$5\% & 0.53$\pm$0.04
      & 52$\pm$7\% & 0.36$\pm$0.06 \\
    Ours
      & \textbf{88$\bm{\pm}$3\%} & \textbf{0.78$\bm{\pm}$0.03}
      & \textbf{88$\bm{\pm}$5\%} & \textbf{0.65$\bm{\pm}$0.05}
      & \textbf{70$\bm{\pm}$6\%} & \textbf{0.46$\bm{\pm}$0.05} \\
    \bottomrule
  \end{tabular}}
\end{table}

\begin{figure*}[t]
    \centering
    \includegraphics[width=\linewidth]{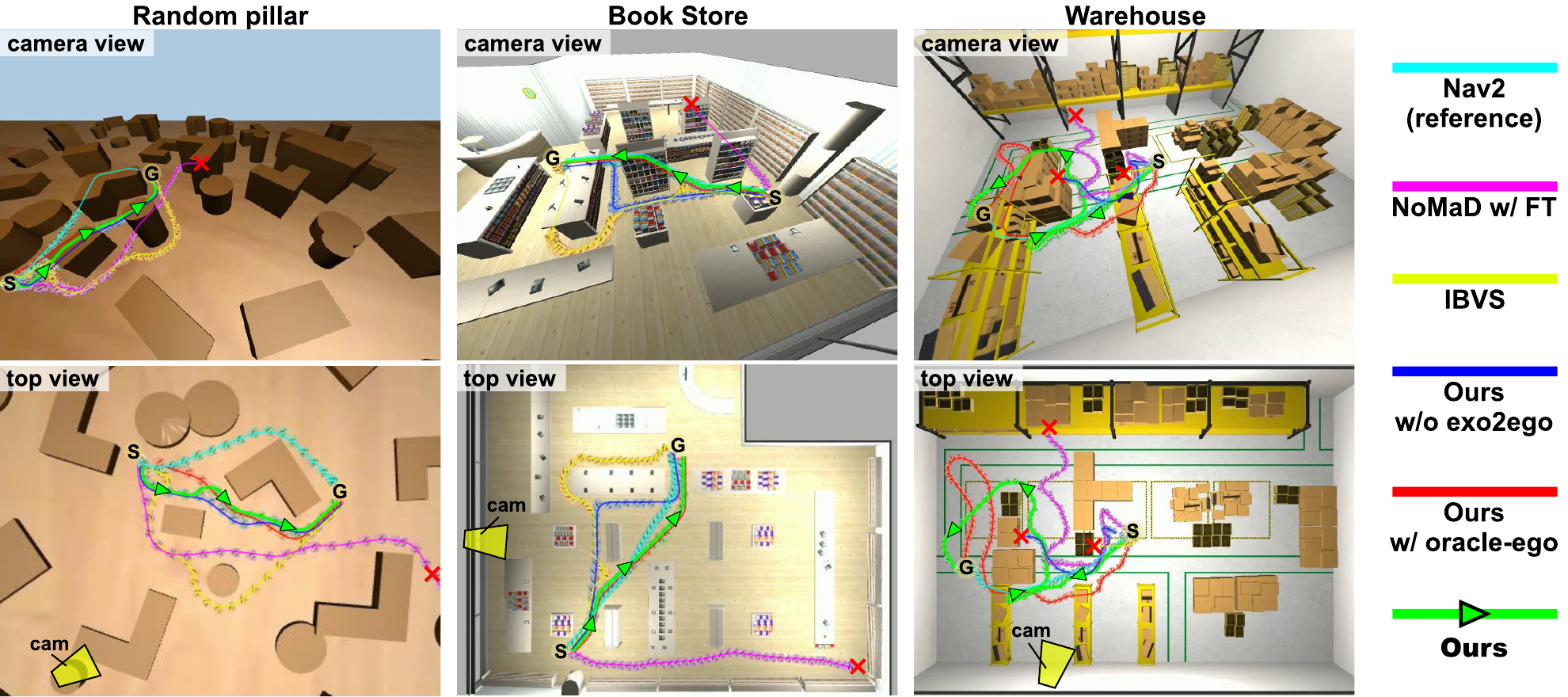}
    \caption{Results in the simulation experiments. Each column is one environment, with the remote camera view on top and a top view of the same trials below. \textsf{S} and \textsf{G} mark the start and the goal, and a cross marks where a run failed.}
    \label{fig:sim-world-result}
\end{figure*}

\begin{figure}[t]
  \centering
  \includegraphics[width=\linewidth]{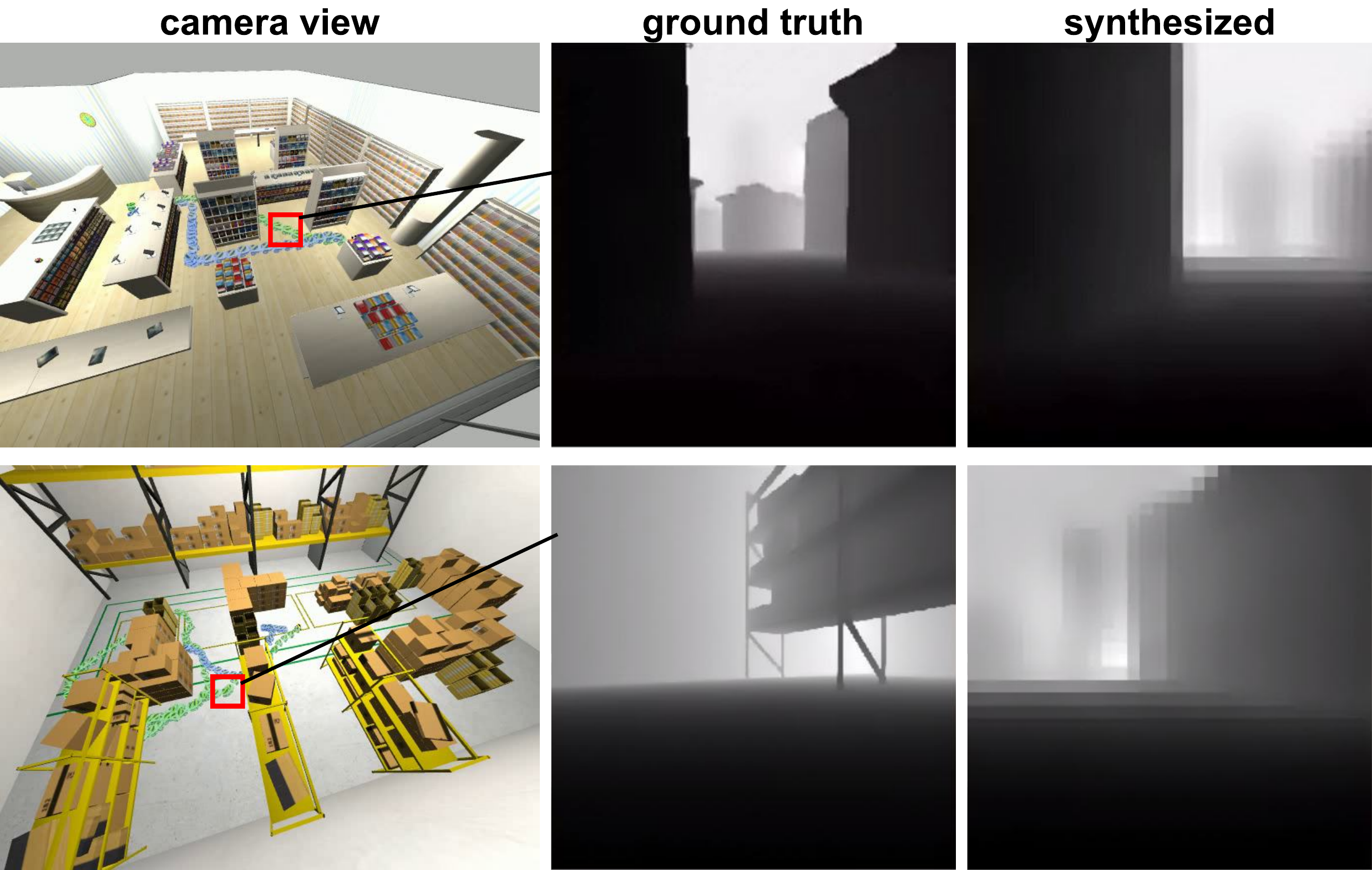}
  \caption{Egocentric depth synthesized by the exo2ego module in \mytag{Book Store} (top) and \mytag{Warehouse} (bottom).}
  \label{fig:exo2ego_images}
\end{figure}

Table~\ref{tab:sim_results} reports SR and SPL in simulation, together with their 95\% confidence intervals. ReVNM achieved the highest performance in all three environments, obtaining an SR\,/\,SPL of 88\%\,/\,0.78 in \mytag{Random Pillar}, 88\%\,/\,0.65 in \mytag{Book Store}, and 70\%\,/\,0.46 in \mytag{Warehouse}. Compared with the IBVS baseline, ReVNM improved success rate by 10, 13, and 18 percentage points, respectively.

NoMaD performed substantially worse under this task setting, with success rates below 40\% even after fine-tuning. Figure~\ref{fig:sim-world-result} illustrates a typical failure mode: after the robot deviates from its pre-recorded reference trajectory, its current observation no longer matches the images stored in the topological map, making recovery difficult.

The IBVS baseline performed considerably better than NoMaD but degraded as the scene geometry became more challenging. Its success rate decreased from 78\% in \mytag{Random Pillar} to 75\% in \mytag{Book Store} and 52\% in \mytag{Warehouse}, while SPL decreased from 0.61 to 0.53 and 0.36. Because IBVS performs global planning directly in image space, its obstacle-clearance margin is represented in pixels. Under an oblique camera, however, the physical distance corresponding to a fixed image-space margin varies across the image because of perspective projection. As a result, the planner may either generate paths with insufficient physical clearance or reject physically traversable gaps. Addressing this issue would require an explicit image-to-ground metric mapping, such as camera or ground-plane calibration, which is outside the calibration-free setting considered here.

\subsubsection{Ablation Study}
\label{sssec:result_ablation}
We ablate two components of ReVNM in Table~\ref{tab:ablation}: the synthesized egocentric observation and the DAgger-inspired recovery data.

\paragraph{Effect of Egocentric View}

We first analyze the contribution of the synthesized egocentric observation. In \mytag{Random Pillar}, ReVNM was comparable to the variant without exo2ego and slightly lower in SR (88\% vs.\ 91\%) and SPL (0.78 vs.\ 0.79). In the more cluttered \mytag{Book Store} and \mytag{Warehouse} environments, however, the benefit of exo2ego became substantial. In \mytag{Warehouse}, removing exo2ego reduced SR from 70\% to 26\% and SPL from 0.46 to 0.17.

These results suggest that exo2ego is particularly useful when the remote view does not sufficiently resolve the robot's local free space. In \mytag{Random Pillar}, obstacles are relatively small and isolated, so the exocentric observation often already provides sufficient local geometric information. In this case, the synthesized observation contributes little and may occasionally introduce prediction noise. In contrast, large shelves and walls in \mytag{Book Store} and \mytag{Warehouse} frequently occlude the space immediately ahead of the robot. The synthesized egocentric depth provides the policy with an estimate of this locally occluded geometry, allowing it to select more appropriate detours.

Figure~\ref{fig:sim-world-result} shows the corresponding trajectories, and Fig.~\ref{fig:exo2ego_images} shows examples of the synthesized egocentric depth. In \mytag{Book Store}, exo2ego predicts the shelf immediately in front of the robot, enabling ReVNM to follow a route close to the Nav2 reference. In \mytag{Warehouse}, where dead ends can place the robot behind large obstacles, exo2ego predicts occupied space ahead and encourages the robot to detour rather than continue forward and stall.

ReVNM also approached the performance of the oracle-ego variant, indicating that exo2ego recovers a substantial fraction of the navigation benefit provided by a true onboard egocentric depth observation, although a noticeable gap remains in the most challenging environments.

\paragraph{Effect of the Recovery Data}
\label{subsec:dagger-effect}

We next evaluate the effect of the DAgger-inspired recovery data. As listed in Table~\ref{tab:ablation}, adding recovery trajectories improved both SR and SPL across all evaluated variants and environments, with the largest gains in cluttered scenes. For example, the success rate of ReVNM in \mytag{Warehouse} increased from 43\% to 70\%, while that of ReVNM without exo2ego in \mytag{Book Store} increased from 44\% to 84\%.

\begin{figure}[t]
  \centering
  \includegraphics[width=\linewidth]{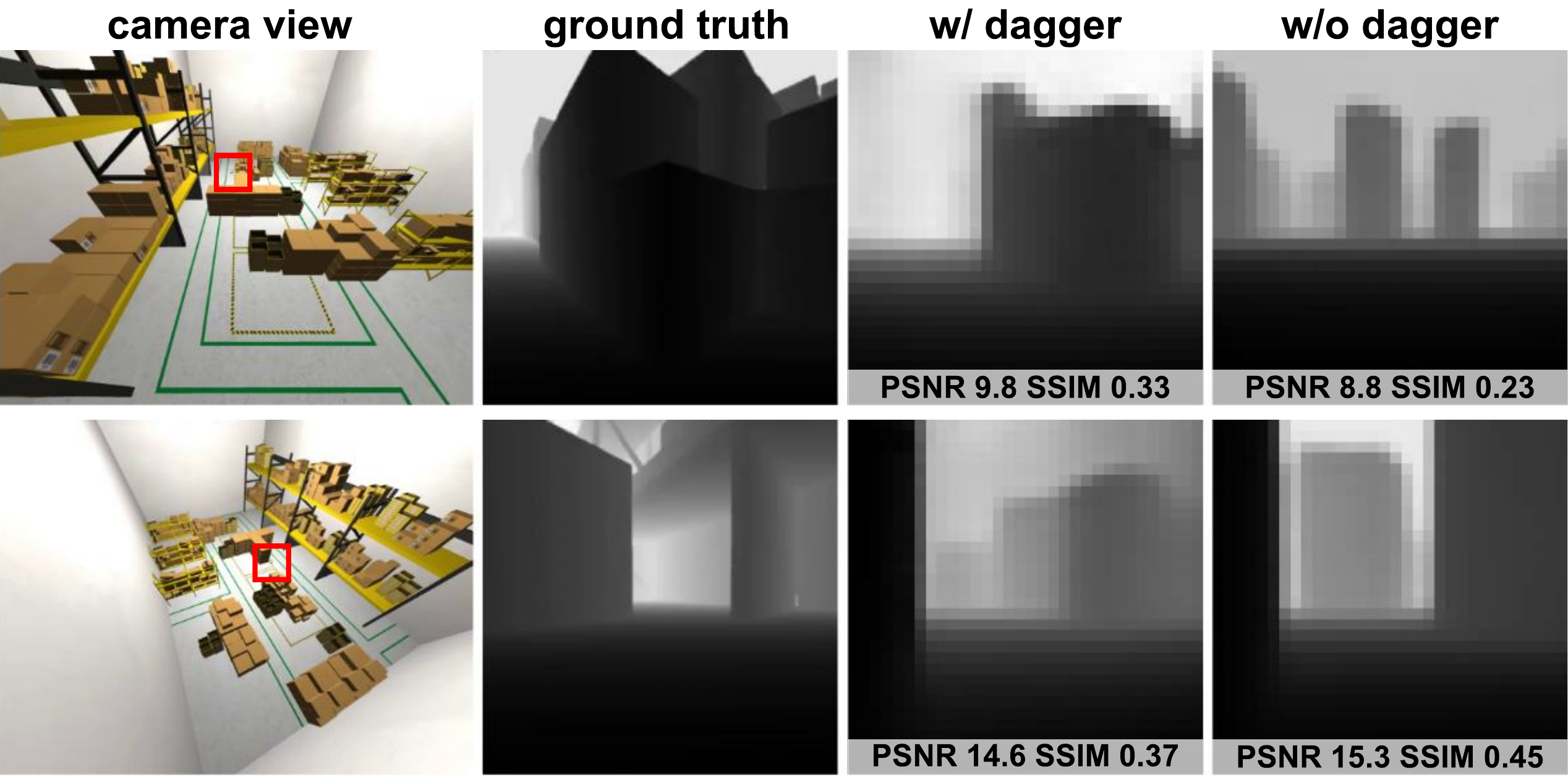}
  \caption{Synthesized ego depth with and without the DAgger-based recovery data. Only w/ DAgger reproduces the obstacle in front of the robot (top), while in open space the two are comparable (bottom).}
  \label{fig:exo2ego_dagger}
\end{figure}

The same trend is reflected in the quality of the synthesized egocentric depth (Table~\ref{tab:exo2ego_quality}). Recovery-data training improved PSNR and FID in all three environments. SSIM improved slightly in \mytag{Random Pillar} but decreased in \mytag{Book Store} and \mytag{Warehouse}. Qualitatively, Fig.~\ref{fig:exo2ego_dagger} shows that recovery data are particularly important when the robot is close to an obstacle: without them, exo2ego rarely sees such states during training and may fail to represent the obstacle immediately ahead.

This behavior arises because nominal Nav2 demonstrations maintain relatively large obstacle clearance. Consequently, training only on expert trajectories creates a distribution shift once the learned policy enters near-obstacle states during closed-loop execution. The recovery-data aggregation procedure explicitly adds such policy-induced states to the training distribution, improving both waypoint prediction and exo2ego prediction in these challenging configurations.

\begin{table}[t]
    \centering
    \caption{Ablation study.}
    \label{tab:ablation}
    \setlength{\tabcolsep}{2.2pt}
    \renewcommand{\arraystretch}{0.95}
    \footnotesize
    \resizebox{\columnwidth}{!}{%
    \begin{tabular}{lccccccc}
      \toprule
      \multirow{2}{*}{Method} & \multirow{2}{*}{DAgger}
        & \multicolumn{2}{c}{\mytag{Random Pillar}}
        & \multicolumn{2}{c}{\mytag{Book Store}}
        & \multicolumn{2}{c}{\mytag{Warehouse}} \\
      \cmidrule(lr){3-4} \cmidrule(lr){5-6} \cmidrule(lr){7-8}
        & & SR\higherbetter & SPL\higherbetter & SR\higherbetter & SPL\higherbetter & SR\higherbetter & SPL\higherbetter \\
      \midrule
      \multirow{2}{*}{Ours}
        & \checkmark & \textbf{88$\bm{\pm}$3\%} & \textbf{0.78$\bm{\pm}$0.03} & \textbf{88$\bm{\pm}$5\%} & \textbf{0.65$\bm{\pm}$0.05} & \textbf{70$\bm{\pm}$6\%} & \textbf{0.46$\bm{\pm}$0.05} \\
        &            & 87$\pm$4\% & 0.69$\pm$0.04 & 74$\pm$6\% & 0.59$\pm$0.06 & 43$\pm$6\% & 0.29$\pm$0.05 \\
      \midrule
      \multirow{2}{*}{Ours w/o exo2ego}
        & \checkmark & \textbf{91$\bm{\pm}$3\%} & \textbf{0.79$\bm{\pm}$0.03} & \textbf{84$\bm{\pm}$5\%} & \textbf{0.61$\bm{\pm}$0.05} & \textbf{26$\bm{\pm}$8\%} & \textbf{0.17$\bm{\pm}$0.06} \\
        &            & 78$\pm$6\% & 0.64$\pm$0.06 & 44$\pm$7\% & 0.33$\pm$0.05 & 12$\pm$4\% & 0.07$\pm$0.03 \\
      \midrule
      \multirow{2}{*}{Ours w/ oracle-ego}
        & \checkmark & \textbf{99$\bm{\pm}$1\%} & \textbf{0.89$\bm{\pm}$0.01} & \textbf{90$\bm{\pm}$4\%} & \textbf{0.78$\bm{\pm}$0.04} & \textbf{83$\bm{\pm}$5\%} & \textbf{0.58$\bm{\pm}$0.05} \\
        &            & 74$\pm$6\% & 0.54$\pm$0.05 & 84$\pm$4\% & 0.64$\pm$0.04 & 71$\pm$6\% & 0.52$\pm$0.06 \\
      \bottomrule
    \end{tabular}}
\end{table}

\begin{table}[t]
    \centering
    \caption{Quality of the synthesized ego depth.}
    \label{tab:exo2ego_quality}
    \setlength{\tabcolsep}{2.2pt}
    \renewcommand{\arraystretch}{0.95}
    \footnotesize
    \resizebox{\columnwidth}{!}{%
    \begin{tabular}{lccccccccc}
      \toprule
      \multirow{2}{*}{exo2ego}
        & \multicolumn{3}{c}{\mytag{Random Pillar}}
        & \multicolumn{3}{c}{\mytag{Book Store}}
        & \multicolumn{3}{c}{\mytag{Warehouse}} \\
      \cmidrule(lr){2-4} \cmidrule(lr){5-7} \cmidrule(lr){8-10}
        & PSNR\higherbetter & SSIM\higherbetter & FID\lowerbetter
        & PSNR\higherbetter & SSIM\higherbetter & FID\lowerbetter
        & PSNR\higherbetter & SSIM\higherbetter & FID\lowerbetter \\
      \midrule
      w/ DAgger
        & \textbf{15.2} & \textbf{0.611} & \textbf{14.5}
        & \textbf{13.1} & 0.454 & \textbf{24.6}
        & \textbf{13.8} & 0.480 & \textbf{23.7} \\
      w/o DAgger
        & 14.4 & 0.601 & 17.3
        & 12.7 & \textbf{0.494} & 25.9
        & 12.5 & \textbf{0.488} & 26.0 \\
      \bottomrule
    \end{tabular}}
\end{table}

\subsection{Navigation Performance in the Real World}


Table~\ref{tab:success_rates} summarizes the real-world navigation results. ReVNM transferred from simulation to a different robot and unseen office configurations without fine-tuning, achieving a 76\% success rate in \mytag{Forest} and succeeding in all five \mytag{Wall} trials. The corresponding success rates of the IBVS baseline were 12\% and 0\%, respectively.

\begin{figure*}[t]
  \centering
  \includegraphics[width=\linewidth]{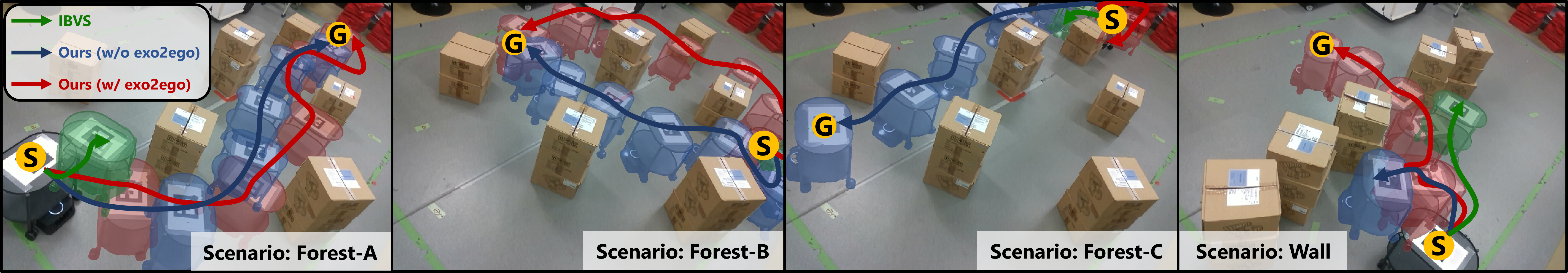}
  \caption{Results in the real-world experiments, overlaid on the remote camera view.}
  \label{fig:real-world-result}
\end{figure*}

Consistent with the simulation results, the contribution of exo2ego depended on the degree of local occlusion. In \mytag{Forest}, where the small obstacles were generally visible from the remote camera, the variant without exo2ego achieved a higher success rate than ReVNM (92\% vs.\ 76\%). In \mytag{Wall}, however, removing exo2ego reduced the success rate from 100\% to 40\%. Figure~\ref{fig:real-world-result} shows that the exocentric-only policy frequently failed to account for the wall directly ahead of the robot, whereas ReVNM used the synthesized egocentric observation to detour and identify the opening.

The IBVS degraded more severely in the real environment. Its image-space obstacle margins provided insufficient clearance for the robot footprint, while increasing the margin caused narrow traversable gaps to disappear. In addition, SAM3 occasionally segmented fine floor texture as obstacle boundaries, fragmenting the estimated traversable region.

These experiments are intended as a demonstration of sim-to-real transfer rather than a comprehensive evaluation across real-world environments. Nevertheless, they exhibit the same qualitative trend observed in simulation: synthesized egocentric information is most useful when the remote observation alone does not sufficiently resolve the local obstacle geometry.

\begin{table}[t]
    \centering
    \caption{Navigation performance in the real-world.}
    \label{tab:success_rates}
    \begin{tabular}{lcc}
      \toprule
      \multirow{2}{*}{Method}
        & \mytag{Forest}
        & \mytag{Wall} \\
      \cmidrule(lr){2-2} \cmidrule(lr){3-3}
        & SR\higherbetter & SR\higherbetter \\
      \midrule
      IBVS             & 12\% (3/25)           & 0\% (0/5)             \\
      Ours w/o exo2ego & \textbf{92\%} (23/25) & 40\% (2/5)            \\
      Ours             & 76\% (19/25)          & \textbf{100\%} (5/5)  \\
      \bottomrule
    \end{tabular}
\end{table}

\section{CONCLUSION}
\label{sec:conclusion}



We presented ReVNM, a learning-based visual navigation approach that enables a mobile robot to navigate using observations from a fixed remote camera. ReVNM predicts robot-centric waypoints from the remote depth image, the robot's image-space pose, and an egocentric depth observation predicted by an exo2ego diffusion module. The system is trained entirely in simulation using procedurally randomized obstacle layouts and camera viewpoints together with DAgger-inspired recovery-data aggregation.

Across three simulated environments, ReVNM outperformed the evaluated image-space remote-navigation and egocentric-VNM baselines in both navigation success and path efficiency. The model also transferred without fine-tuning to a different real robot in two office navigation scenarios. Our ablations show that the synthesized egocentric observation is particularly beneficial when obstacles occlude the robot's local surroundings from the remote camera, whereas it provides less benefit when the exocentric view already resolves the nearby free space. Recovery-data aggregation further improves robustness by exposing the model to near-failure states that are rarely encountered in nominal expert demonstrations.

\textbf{Limitations and Future Work.}
ReVNM currently assumes that the robot can be localized in the remote-camera image and that sufficient visual coverage of the workspace is maintained. In our real-world experiments, the robot pose is obtained using an AprilTag; practical deployment will require robust marker-free pose estimation. A multi-camera extension with camera handover could address limited field-of-view coverage. In addition, the synthesized egocentric depth can be inaccurate or overly conservative. Future work could fuse the exocentric and synthesized observations in an uncertainty-aware manner.



\balance


\bibliographystyle{IEEEtran}
\bibliography{IEEEabrv, venues_abbrev, reference}

\end{document}